\documentclass{article}

\usepackage{amsmath,amsfonts,bm}

\def\eqref#1{equation~\ref{#1}}
\def\1{\bm{1}}

\def\rvh{{\mathbf{h}}}

\def\rvx{{\mathbf{x}}}
\def\rvy{{\mathbf{y}}}

\def\mH{{\bm{H}}}

\DeclareMathAlphabet{\mathsfit}{\encodingdefault}{\sfdefault}{m}{sl}
\SetMathAlphabet{\mathsfit}{bold}{\encodingdefault}{\sfdefault}{bx}{n}

\newcommand{\R}{\mathbb{R}}

\usepackage{xspace}
\makeatletter
\DeclareRobustCommand\onedot{\futurelet\@let@token\@onedot}
\def\@onedot{\ifx\@let@token.\else.\null\fi\xspace}

\def\eg{\emph{e.g}\onedot}

\makeatother

\usepackage{microtype}
\usepackage{graphicx}
\usepackage{subfigure}
\usepackage{booktabs} % for professional tables

\usepackage{hyperref}

\usepackage[accepted]{icml2025}

\usepackage{amsmath}
\usepackage{amssymb}
\usepackage{mathtools}
\usepackage{amsthm}

\usepackage[capitalize,noabbrev]{cleveref}

\usepackage{url}
\usepackage{xcolor}

\usepackage{listings}

\definecolor{codegreen}{rgb}{0,0.6,0}
\definecolor{codegray}{rgb}{0.5,0.5,0.5}
\definecolor{codepurple}{rgb}{0.58,0,0.82}
\definecolor{backcolour}{rgb}{0.95,0.95,0.92}

\lstdefinestyle{mystyle}{
    backgroundcolor=\color{backcolour},   
    commentstyle=\color{codegreen},
    keywordstyle=\color{magenta},
    numberstyle=\tiny\color{codegray},
    stringstyle=\color{codepurple},
    basicstyle=\ttfamily\footnotesize,
    breakatwhitespace=false,         
    breaklines=true,                 
    captionpos=b,                    
    keepspaces=true,                 
    numbers=right,                    
    numbersep=5pt,                  
    showspaces=false,                
    showstringspaces=false,
    showtabs=false,                  
    tabsize=2
}

\usepackage[capitalize]{cleveref}
\Crefname{equation}{Eq.}{Eqs.}
\Crefname{equations}{Eq.}{Eqs.}
\Crefname{figure}{Fig.}{Figs.}
\Crefname{figures}{Figs.}{Figs.}
\Crefname{section}{Sec.}{Secs.}
\Crefname{sections}{Secs.}{Secs.}
\Crefname{table}{Tab.}{Tabs.}
\Crefname{tables}{Tabs.}{Tabs.}
\Crefname{appendix}{App.}{Apps.}
\Crefname{appendices}{App.}{Apps.}
\Crefname{algorithm}{Alg.}{Algs.}
\Crefname{algorithms}{Alg.}{Algs.}

\theoremstyle{plain}

\theoremstyle{definition}

\theoremstyle{remark}

\usepackage[disable,textsize=tiny]{todonotes}
\icmltitlerunning{PSC: Posterior Sampling-Based Compression}

\begin{document}

\twocolumn[
\icmltitle{PSC: Posterior Sampling-Based Compression}



\icmlsetsymbol{equal}{*}

\begin{icmlauthorlist}
\icmlauthor{Noam Elata}{ece}
\icmlauthor{Tomer Michaeli}{ece}
\icmlauthor{Michael Elad}{cs}
\end{icmlauthorlist}

\icmlaffiliation{ece}{Department of Electric and Computer Engineering, Technion - Israel Institute of Technology, Haifa, Israel}
\icmlaffiliation{cs}{Department of Computer Science, Technion - Israel Institute of Technology, Haifa, Israel}

\icmlcorrespondingauthor{Noam Elata}{noamelata@campus.technion.ac.il}

\icmlkeywords{Machine Learning, ICML, Image Compression, Diffusion Models, Posterior Sampling, Zero-Shot, pre-trained}

\vskip 0.3in
]



\printAffiliationsAndNotice{}  % leave blank if no need to mention equal contribution

\begin{abstract}
Diffusion models have transformed the landscape of image generation and now show remarkable potential for image compression. Most of the recent diffusion-based compression methods require training and are tailored for a specific bit-rate. In this work, we propose Posterior Sampling-based Compression (PSC) -- a zero-shot compression method that leverages a pre-trained diffusion model as its sole neural network component, thus enabling the use of diverse, publicly available models without additional training. Our approach is inspired by transform coding methods, which encode the image in some pre-chosen transform domain. However, PSC constructs a transform that is adaptive to the image. This is done by employing a zero-shot diffusion-based posterior sampler so as to progressively construct the rows of the transform matrix. Each new chunk of rows is chosen to reduce the uncertainty about the image given the quantized measurements collected thus far. Importantly, the same adaptive scheme can be replicated at the decoder, thus avoiding the need to encode the transform itself. We demonstrate that even with basic quantization and entropy coding, PSC's performance is comparable to established training-based methods in terms of rate, distortion, and perceptual quality. This is while providing greater flexibility, allowing to choose at inference time any desired rate or distortion.
\end{abstract}

\section{Introduction}

Diffusion models excel at generating high-quality images~\cite{ho2020denoising, sohl2015deep, song2020score, dhariwal2021diffusion, vahdat2021score, latent_diffusion}. Their versatility has enabled solutions for diverse tasks, including inverse problems~\cite{saharia2021image, saharia2022palette, chung2023diffusion, kawar2021snips, kawar2022denoising, song2023pseudoinverse}, image editing~\cite{meng2021sdedit, brooks2022instructpix2pix, kawar2023imagic, huberman2023edit}, and uncertainty quantification~\cite{belhasin2023principal, horwitz2022conffusion}. Conveniently, many of these applications can utilize pre-trained diffusion models without requiring task-specific training.

Image compression is fundamental for efficient storage and transmission of visual data, attracting significant research attention over decades. Effective compression schemes preserve essential image information while discarding less critical components, establishing a lossy compression paradigm that balances image quality against file size.
Traditional compression methods like JPEG~\cite{wallace1991jpeg} and JPEG2000~\cite{skodras2001jpeg} achieve this through fixed whitening transforms and coefficient quantization, allocating bits based on the coefficients' importance and applying entropy coding for additional lossless compression. 
More recently, neural compression methods have demonstrated improved performance over their classical counterparts, by incorporating quantization and entropy-coding directly into their training objectives~\cite{balle2018variational, minnen2018joint, cheng2020image, balle2016end, theis2017lossy, toderici2015variable}. In this context, deep generative models, such as GANs~\cite{mentzer2020high} or diffusion models~\cite{yang2024lossy}, can improve the perceptual quality of decompressed images. 
While some approaches use generative modeling for post-hoc restoration of existing compression algorithms~\cite{bahat2021s,kawar2022jpeg, saharia2022palette, man2023high, song2023pseudoinverse, ghouse2023residual, xu2024idempotence}, their performance gains remain limited as they do not modify the encoder. Alternative methods that integrate neural compression with diffusion model-based decoding~\cite{careil2023towards, yang2024lossy, relic2024lossy} achieve impressive results but require rate-specific diffusion model training for a single modality, leading to reduced flexibility.

\begin{figure*}[t]
\vskip 0.2in
\begin{center}
  \centerline{\includegraphics[width=0.88\textwidth]{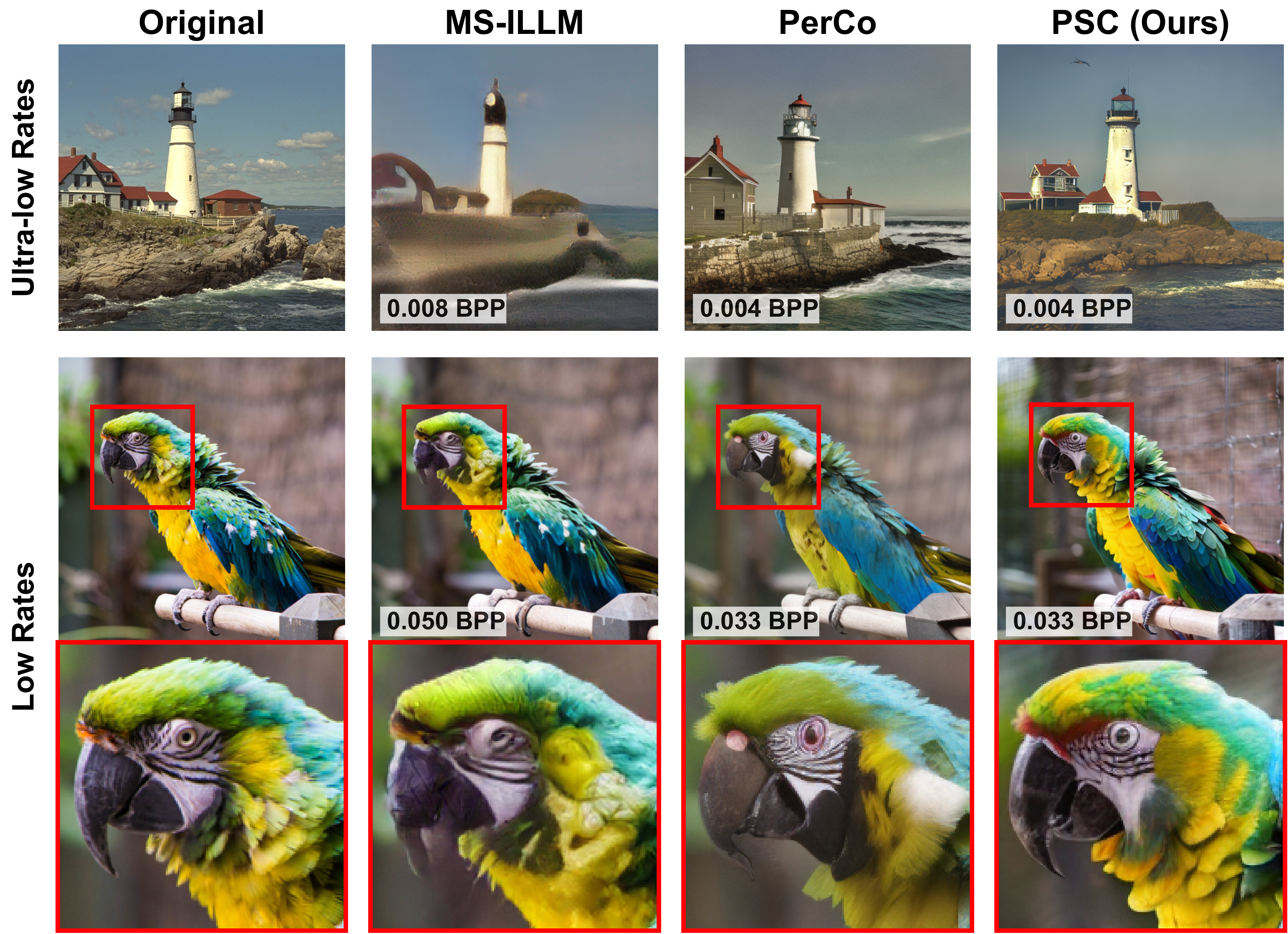}}
  \caption{\textbf{Images compressed with latent-PSC at low bit-rates.} Latent-PSC leverages pre-trained diffusion models to deliver high perceptual quality at any compression rate, indicated by the bits-per-pixel (BPP) for each decompressed result. Top: Example at ultra-low rates demonstrates PSC maintains high image quality while preserving the original image composition. Bottom: Example at low rates with zoomed detail highlights PSC's capacity to maintain fine details, adapting seamlessly to any compression rate without training.}
    \label{fig:latent-examples}
\end{center}
\vskip -0.2in
\end{figure*}

In this paper, we introduce a lossy compression method that uses a pre-trained diffusion model. This allows leveraging publicly available models without requiring additional training, and thus enables lossy compression of any data type on which a diffusion model has been trained. Our paradigm is very different from recent neural compression literature, and can be considered an adaptive version of classical transform coding techniques. Specifically, our work draws inspiration from AdaSense~\cite{elata2024adaptive}, which utilizes a pre-trained diffusion-model to identify an image-specific linear transform, from which the image can be reconstructed with near-minimal error. Here, we propose to use the same approach but for constructing an \emph{image-specific compression transform}. A key challenge in doing so, is that in addition to encoding the transformed image, this approach seemingly requires encoding also the transform itself, which is completely impractical. 
However, our key observation is that the communication of the transform can be bypassed altogether. This allows our method to perform well at extremely low bit rates, as we illustrate in Fig.~\ref{fig:latent-examples}. 

Our compression scheme, which we coin Posterior Sampling-based Compression (PSC), uses a diffusion-based posterior sampler to progressively transform and quantize the given image. In each step, the next rows of the transform matrix are chosen in a way that minimizes the uncertainty that remains about the image, given the code constructed so far. The decoder replicates these calculations (fixing the same seed) such that both the encoder and the decoder reproduce exactly the same image-adaptive transform, eliminating the need for transmitting side-information beyond the quantized coefficients of the image in the transform domain.

PSC can be applied at any rate or distortion level prescribed by the user at inference time, and enables decompression at high perceptual quality. 
We evaluate PSC's effectiveness against established compression methods on the ImageNet dataset~\cite{imagenet}, comparing both distortion and image quality.
We also explore the compression of general images using state-of-the-art (SOTA) text-to-image latent diffusion models~\cite{latent_diffusion}, incorporating textual descriptions within the compressed representation. We evaluate this latent-PSC variant against high-perceptual-quality compression algorithms~\cite{careil2023towards, muckley2023illm} on the CLIC~\cite{CLIC2020} and DIV2K~\cite{div2k} datasets. 
Our experiments demonstrate that PSC is competitive with established methods, despite using a pre-trained diffusion model as its sole neural network component. While our preliminary implementation employs simplified quantization, lacks tailored entropy coding and has high computational complexity, the results underscore the potential of our approach for flexible compression applications. 
Future advances in diffusion-based posterior samplers, combined with our training-free compression framework, have the potential to yield significant improvements in compression of images as well as other signals of interest.

\begin{algorithm}[b]
\caption{A single iterative step of row selection \newline Denoted as $\text{SelectNewRows}\left( \mH_{0:k}, \rvy_{0:k}, r \right)$}
\label{alg:adasense-step}
 \begin{algorithmic}[1]
    \REQUIRE Previous sensing rows $\mH_{0:k}$, corresponding measurements $\rvy_{0:k}$, number of new measurements $r$
    \STATE $\{\rvx_i\}_{i=1}^s \sim p(\rvx|\rvy_{0:k},\mH_{0:k})$  
    % \Comment{\textcolor{gray}{generate $s$ posterior samples}}
    \STATE $\{\rvx_i\}_{i=1}^s \gets \{\rvx_i - \frac{1}{s} \sum_{j=1}^s \rvx_j \}_{i=1}^s$ 
    % \Comment{\textcolor{gray}{center samples}}
    \STATE $\Tilde{\mH} \gets \text{Append top $r$ singular vectors of }\begin{pmatrix}\rvx_1 , \dots , \rvx_s \end{pmatrix}^\top $ 
    
    % \Comment{\textcolor{gray}{select $r$ principal components}}
    \STATE \textbf{return} $\Tilde{\mH}$ 
 \end{algorithmic}
\end{algorithm}

To summarize, PSC offers the following advantages:
% \vspace{-0.1in}
\begin{itemize}
    \item It provides precise instance-level control over rate and distortion. This is achieved by accumulating measurements until reaching the desired rate or distortion level.
    \item For the same compressed code, decoding can aim for either high-quality or low-distortion. This is done by appropriately choosing the final restoration algorithm.
    \item As a zero-shot approach, PSC generalizes to any data type supported by diffusion models, extending beyond image-specific applications.
    \item Our method will inherently benefit from future advances in generative modeling and posterior sampling, without requiring modifications or re-training.
\end{itemize}

%=============================================

\begin{figure}[t!]
\vskip 0.2in
\begin{center}
  \centerline{\includegraphics[width=\columnwidth]{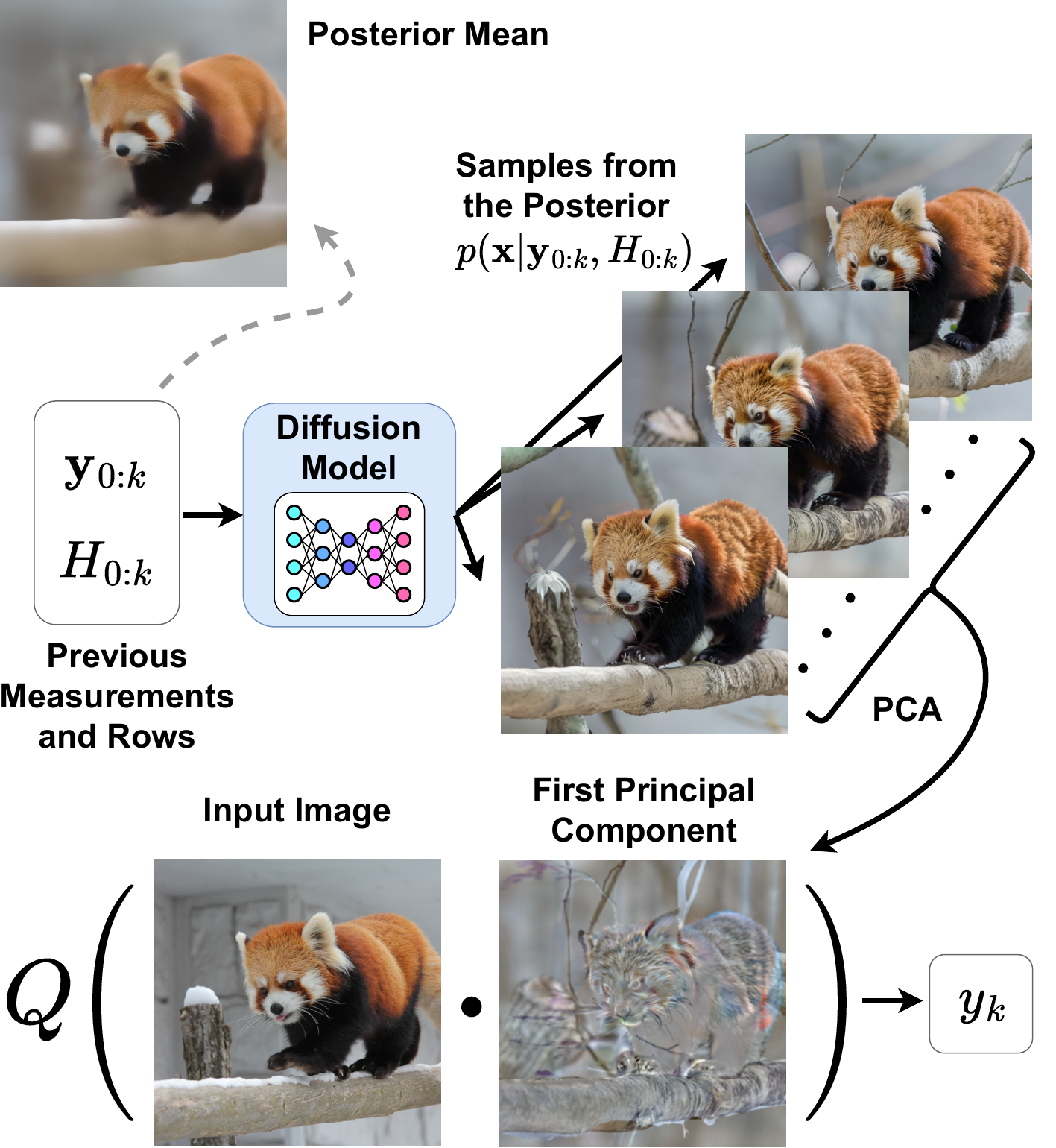}}
  \caption{\textbf{A diagram of how PSC zeros in on the input image by progressively tightening the posterior.} The diagram shows a single step, where the new rows and matching measurements are computed based on the direction of largest uncertainty in the posterior distribution. The posterior mean is shown to visualize the information captured by previous iterations.}
    \label{fig:toy-example}
\end{center}
\vskip -0.2in
\end{figure}

\section{Background}
\label{sec:background}
\subsection{Posterior Sampling for Linear Inverse Problems}
\label{sec:inverse}
Linear inverse problems are tasks in which a signal $\rvx$ needs to be recovered from linear measurements $\rvy = \mH\rvx$, where $\mH\in \R^{d\times D}$ is a known matrix. One approach to obtaining plausible reconstructions, is to draw samples from the posterior distribution $p(\rvx|\rvy)$. Recent research demonstrates that pre-trained diffusion models, which were originally trained to sample from the prior distribution $p(\rvx)$, can be adopted in a zero-shot manner to approximately sample from the posterior distribution~\cite{kawar2022denoising, chung2023diffusion, song2023pseudoinverse}. This can be done for any vector of measurements $\rvy$ and any measurement matrix $\mH$, which are provided to the sampler at inference time.

\subsection{Adaptive Compressed Sensing}
In several important problems, such as single pixel cameras and MRI, there is flexibility in designing the operator $\mH$. This gives rise to a compressed sensing problem~\cite{donoho2006compressed}, in which the question is which matrix $\mH$ (of predefined dimensions) allows reconstructing the signals of interest with minimal error. 
Recently, \citet{elata2024adaptive} introduced AdaSense, a method that utilizes posterior samplers to progressively construct the sensing matrix $\mH$ so as to optimally represent the input image. 

The algorithm works as follows. At stage $k$, we have the currently held\footnote{In our notations, the subscript $\{0:k\}$ implies that $k$ elements are available, from index $0$ to index $k-1$. } matrix $\mH_{0:k}$ and measurements $\rvy_{0:k} = \mH_{0:k}\rvx$. The selection of the next row is done by generating samples from the posterior $p(\rvx|\rvy_{0:k},\mH_{0:k})$ using any posterior sampler and applying principal component analysis (PCA) to identify the principal direction of the uncertainty that remains in $\rvx$ given the current $\rvy_{0:k}$.  This direction is chosen as the next row in $\mH$, which is used to acquire a new measurement of $\rvx$. More generally, instead of selecting one new measurement at a time, it is possible to add $r$ new measurements in each iteration\footnote{This algorithm presents a strategy of choosing the $r$ leading eigenvectors of the PCA at every stage instead of a single one, getting a substantial speedup in the measurements' collection process at a minimal cost to adaptability.}. This selection of the new rows of $\mH$ is optimal in the sense that it allows achieving the minimal possible mean squared error (MSE) that can be obtained with a linear decoder (even though a more sophisticated reconstruction method is eventually used). 

A single iteration of the method is detailed in \Cref{alg:adasense-step}. The process repeats until $\mH$ reaches the desired dimensions. 
It is important to note that AdaSense produces an \emph{image-specific sensing matrix $\mH$} and corresponding measurements~$\rvy$. These can be used for obtaining a candidate reconstruction $\hat{\rvx}$ by leveraging the final posterior, $p(\rvx|\rvy,\mH)$, where $\mH$ is the final matrix (obtained at the last step). This final reconstruction may employ a different, more accurate (and possibly more computational demanding) posterior sampler, as it executes only once. Please see \cite{elata2024adaptive} for further details.

\begin{algorithm}[t]
\caption{PSC Encoder}
\label{alg:PSC-enc}
\begin{algorithmic}[1]
\REQUIRE Image $\rvx$, number of steps $N$, number of measurements per step $r$.
\STATE \textbf{initialize} $\rvy_{0:0}, \mH_{0:0}$ as an empty vector or matrix
\FOR{$n\in\{0 : N-1\}$}
    \STATE $\mH_{nr:nr+r} \gets \text{SelectNewRows}\left( \mH_{0:nr}, \rvy_{0:nr}, r \right)$ 
    \STATE $\rvy_{0:nr+r} \gets \text{Append} \left[ \rvy_{0:nr}, Q (\mH_{nr:nr+r} \rvx) \right]$ 
    \STATE $\mH_{0:nr+r} \gets \text{Append} \left[ \mH_{0:nr}, \mH_{nr:nr+r} \right]$
\ENDFOR
\STATE \textbf{return} $\text{Entropy Encode}(\rvy_{0:Nr})$ 
\end{algorithmic}
\end{algorithm}

\subsection{Transform Coding}

Image compression algorithms based on the transform coding paradigm, like the widely used JPEG~\cite{wallace1991jpeg}, apply a pre-chosen, fixed and orthonormal\footnote{Having orthogonal rows has two desirable effects -- easy-inversion and a whitening effect. Using a biorthogonal system as in JPEG2000~\cite{skodras2001jpeg} has similar benefits. } transform on the input image, $\rvx\in\R^D$, obtaining its representation coefficients. These coefficients go through a quantization stage, in which portions of the transform coefficients are discarded entirely, and other portions are replaced by finite precision versions, with a bit-allocation that depends on their importance for the image being compressed. As some of the transform coefficients are discarded, this scheme can be effectively described as using a partial transform matrix $\mH\in\R^{d\times D}$ with orthogonal rows, and applying the quantization function $Q(\cdot)$ to each of elements of the remaining measurements $\rvy = \mH\rvx$.
Image compression algorithms include an entropy coding stage that takes the created bit-stream and passes it through a lossless coding block (\eg Huffman coding, arithmetic coding, etc.) for a further gain in the resulting file-size.

The decoder has knowledge of the transform used, $\mH$. Therefore, given the encoded signal, $Q(\rvy)$, the decoder can generate a reconstructed image \eg by using  $\mH^{\dagger} Q(\rvy)$, where $\mH^{\dagger}$ is the Moore–Penrose pseudo-inverse of $\mH$. In practice, more sophisticated restoration methods are often used, some of which make use of neural networks. 

When a compression algorithm is said to be progressive, this means that the elements of $\rvy$ are sorted based on their importance, and transmitted in their quantized form sequentially, enabling a decompression of the image at any stage based on the received coefficients so far. Progressive compression algorithms are highly desirable, since they induce a low latency in decompressing the image. 
Note that the progressive strategy effectively implies that the rows of $\mH$ have been sorted as well based on their importance, as each row gives birth to the corresponding element in $\rvy$. 
Adopting this view, at step $k$ we consider the sorted portions of $\mH$ and $\rvy$, denoted by $\mH_{0:k}\in\R^{k\times D}$ and $\rvy_{0:k} =\mH_{0:k}\rvx \in\R^k$. As the decoder gets $Q(\rvy_{0:k})$, it may produce $\mH_{0:k}^\dagger Q(\rvy_{0:k})$ as a temporary output image. It is important to note that in this paradigm $\mH$ is fixed, and therefore the sorting of its rows is determined only once. This chosen order is used for compressing every image $\rvx$. 

\begin{figure*}[t]
\vskip 0.2in
\begin{center}
  \centerline{\includegraphics[width=0.90\textwidth]{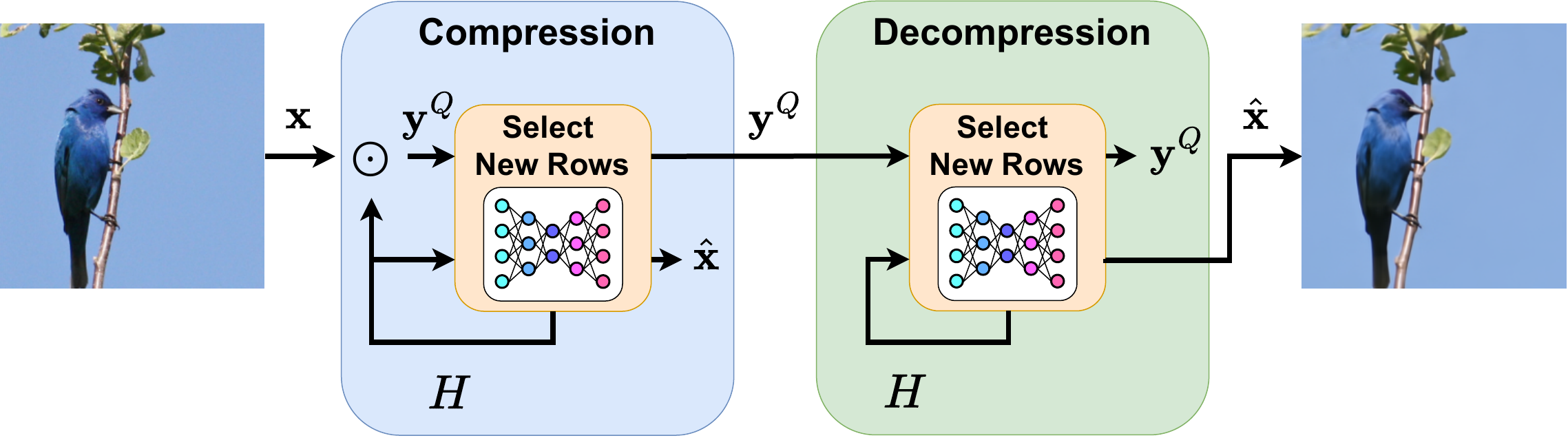}} \caption{\textbf{PSC diagram:} Both encoder and decoder construct an image-specific transform $\mH$ through an adaptive compressed sensing algorithm, progressively adding rows based on posterior sample covariance. The transmission of quantized measurements $\rvy$ ensures identical inputs at each progressive step, while a shared random seed guarantees deterministic outputs on both sides. Together, these factors enable the construction of identical transforms on both sides -- eliminating the need to transmit the transform as side information.}
    \label{fig:diagram}
\end{center}
\vskip -0.2in
\end{figure*}

%=============================================

\section{Method}
\label{sec:method}

\begin{algorithm}[t]
\caption{PSC Decoder}
\label{alg:PSC-dec}
\begin{algorithmic}[1]
\REQUIRE compressed representation $\rvy$, number of steps $N$, number of measurements per step $r$.
\STATE \textbf{initialize} $\rvy_{0:Nr} \gets \text{Entropy Decode}(\rvy)$ 
\STATE \textbf{initialize} $\mH_{0:0}$ as an empty matrix
\FOR{$n\in\{0 : N-1\}$}
    \STATE $\mH_{nr:nr+r} \gets \text{SelectNewRows}\left( \mH_{0:nr}, \rvy_{0:nr}, r \right)$ 
    \STATE $\mH_{0:nr+r} \gets \text{Append} \left[ \mH_{0:nr}, \mH_{nr:nr+r} \right]$
\ENDFOR
\STATE \textbf{return} $\hat{\rvx} = f(\rvy_{0:Nr}, \mH_{0:Nr}) $ 
\end{algorithmic}
\end{algorithm}

In this section, we present how our proposed compression scheme builds upon the framework of transform coding by leveraging the learned prior of diffusion models to determine an \emph{image-specific transform}. By adapting the compressed sensing algorithm detailed in \cref{sec:inverse} to compression, PSC implements a progressive approach to select the most informative partial measurements, effectively reducing reconstruction error by progressively increasing the bit-stream length. We solve the conundrum created by using an image-specific transform, by demonstrating that both the encoder and the decoder can synchronize the transform without any use of side-information.

During each step of the compression phase, our system dynamically estimates the posterior probability distribution $p(\rvx|\rvy_{0:k},\mH_{0:k})$, which is conditioned on the previously extracted $\mH_{0:k}$ and the corresponding quantized partial measurements $\rvy_{0:k}$. This estimation utilizes samples generated by a posterior sampler. Importantly, the zero-shot methods listed in \Cref{sec:inverse} can solve any inverse problem of the form $\rvy = Q(\mH \rvx)$, enabling the utilization of pre-trained diffusion models without training. In practice, we use posterior samplers designed for linear inverse problems (without quantization).  This allows using efficient samplers and leads to sufficiently accurate results. 

The selection of the next row of $\mH$, denoted as $\rvh_k\in\R^{1 \times D}$, is determined by identifying the eigenvector corresponding to the largest eigenvalue of the posterior covariance. This method ensures the projection of $\rvx$ occurs along the most informative direction, maximizing the value of incremental information gathered. The resulting measurement, $y_k = Q(\rvh_k \rvx)$, is then appended to the previous compressed representation $\rvy_{0:k}$ to form $\rvy_{0:k+1}$. This process effectively reduces the uncertainty of candidate images within the posterior distribution $p(\rvx|\rvy_{0:k},\mH_{0:k})$ in a nearly optimal manner. Interestingly, as a by-product of the algorithm, the obtained sensing matrix $\mH$ has orthogonal rows, disentangling the measurements, as expected from a compression algorithm. 
A single step of this process is described in \Cref{fig:toy-example}.

This approach might appear counterintuitive for a transform coding, as it raises questions about the decoder's ability to determine the appropriate transform for image recovery. The naive approach of directly communicating the transform would be impractical, as it would require more bits than a lossless transmission of the image itself (the transform can be any matrix in $\R^{d \times D}$). However, we present an elegant solution that circumvents this challenge through our progressive compression structure, which eliminates the need to communicate the transform entirely. 

More specifically, PSC maintains a synchronized and identical state between encoder and decoder throughout its operation. 
The system relies on an agreed-upon seed, ensuring all random sampling operations produce deterministic and reproducible outputs. PSC initiates both encoder and decoder algorithms from the same empty matrix $\mH_{0:0}$ and empty vector of previous quantized projections $\rvy_{0:0}$. 
Assuming that the previous steps were completed successfully -- i.e. the accumulated matrix $\mH_{0:k}$ and quantized measurements $\rvy_{0:k}$ are identical in both the encoder and decoder, we proceed to construct the next row of $\mH_{0:k}$. 
During this computation, % of this next row, 
all posterior samples $\{\rvx_i\}_{i=1}^s \sim p(\rvx|\rvy_{0:k},\mH_{0:k})$ are identical in both encoder and decoder, ensuring the synchronization of the newly computed row $\rvh_k$. The encoder then evaluates and quantizes the new measurements $y_k = Q(\rvh_k \rvx)$, incorporating them into the compressed representation transmitted to the decoder. Using the compressed representation, the decoder can utilize all measurements $\rvy_{0:k+1}$ directly in subsequent steps without requiring access to the input image. This ensures that the inputs to the next iteration, $\mH_{0:k+1}$ and $\rvy_{0:k+1}$, remain synchronized. This novel approach to synchronized transform reconstruction is illustrated in \Cref{fig:diagram}. The complete procedures for compression and decompression with PSC, including the optimization of selecting $r$ rows from matrix $\mH$ for improved efficiency, are detailed in \Cref{alg:PSC-enc} and \Cref{alg:PSC-dec} respectively.

\begin{figure*}[t]
\vskip 0.2in
\begin{center}
  \centerline{\includegraphics[width=0.92\textwidth]{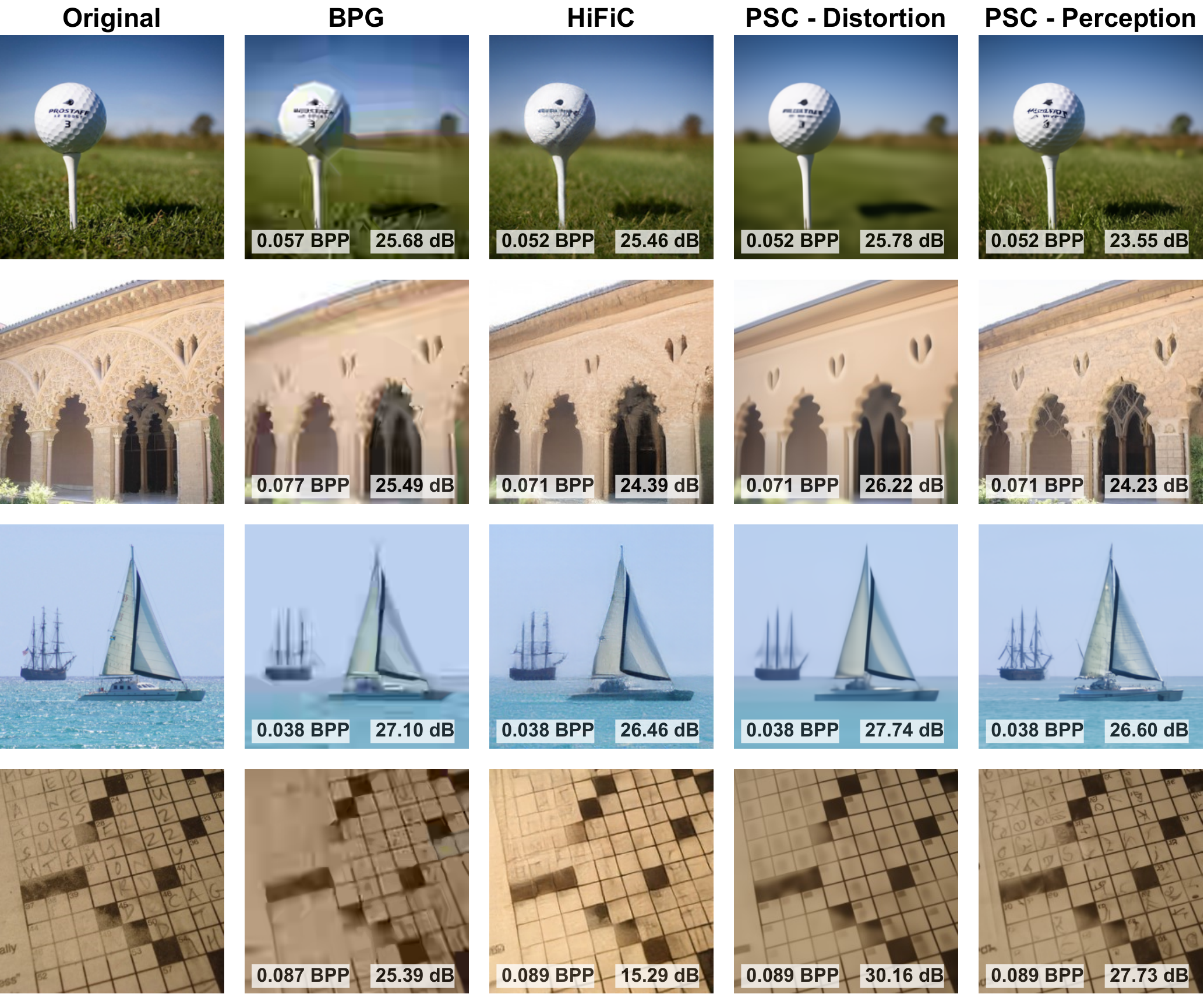}}
  \caption{\textbf{Qualitative examples for compression with PSC, compared to other compression algorithms.} BPP and PSNR are reported per example. Our method can be used for both low-distortion or high perceptual quality using the same compressed representation.}
    \label{fig:imagenet-256}
\end{center}
\vskip -0.2in
\end{figure*}

\begin{figure*}[t]
\vskip 0.2in
  \begin{center}
  \centerline{\includegraphics[width=\textwidth]{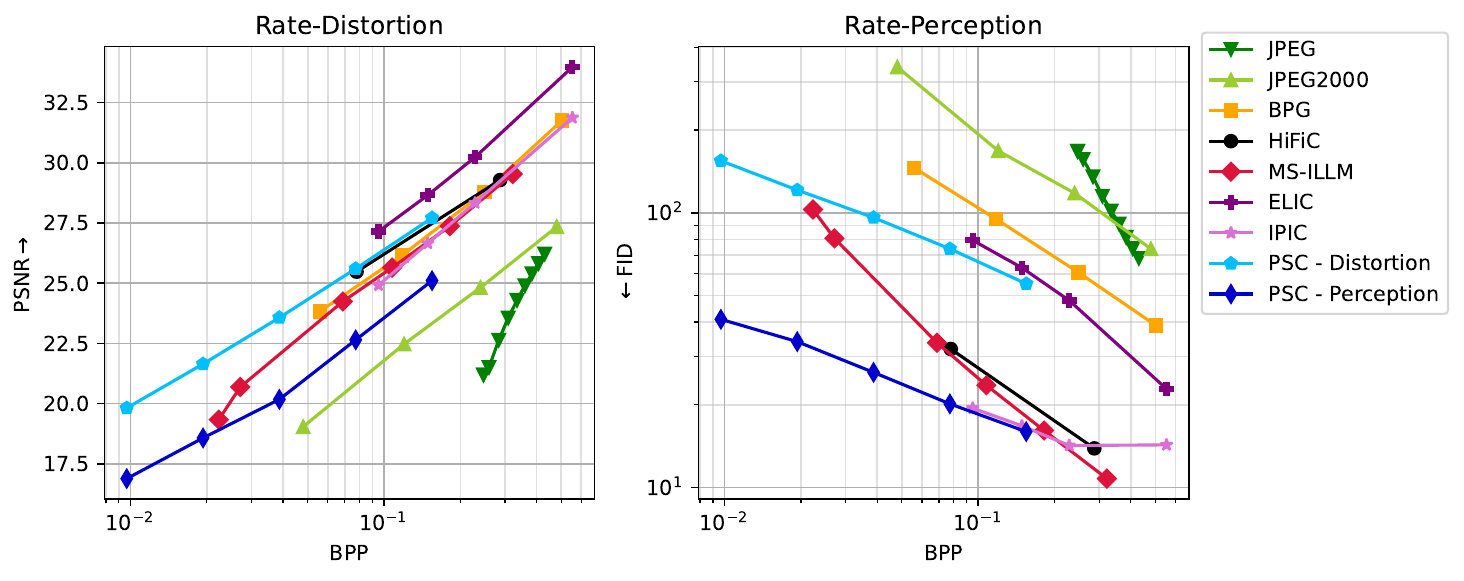}}
  \caption{\textbf{Rate-Distortion (left) and Rate-Perception (right) curves for ImageNet256 compression.} Distortion is measured as average PSNR of images for the same desired rate or specified compression quality, while Perception (photorealism) is measured by FID.}
    \label{fig:imagenet-256-curves}
  \end{center}
  \vskip -0.2in
\end{figure*}

\subsection{Implementation Choices}
We use DDRM~\cite{kawar2022denoising} as a zero-shot posterior sampler for the selection of $\mH$, due to it's relative low computational complexity. Due to the repeated sampling from different posteriors, PSC retains a high computational complexity, requiring approximately 10,000 NFEs for both compression and decompression. Nevertheless, we expect advances in diffusion models
and posterior sampling to significantly expedite future versions.
In our implementation we focus on an unsophisticated quantization approach, reducing the precision of $\rvy$ from float32 to float8~\cite{micikevicius2022fp8}. We employ Range Encoding implemented using~\cite{bamler2022constriction} as an entropy coding on the quantized measurements. The quantization, the posterior sampler and the entropy coding could all be improved, posing promising directions for future work. Finally, after reproducing $\mH$ on the decoder side, PSC may use a different final posterior sampler during decompression, in an attempt to further boost perceptual quality for the very same measurements $\rvy$.

\section{Experiments}

We begin with an evaluation of PSC on $256\times256$ color images from the ImageNet~\cite{imagenet} dataset, using the unconditional diffusion models from~\cite{dhariwal2021diffusion} as an image prior. We compare distortion (PSNR) and bits-per-pixel (BPP) averaged on a subset of validation images, using one image from each of the 1000 classes, following~\cite{pan2021exploiting}. We apply PSC to progressively decode at higher rates, as detailed in \Cref{app:implementation}.

A key advantage of PSC is its ability to prioritize perceptual quality during decompression by changing the final reconstruction algorithm. However, this flexibility comes with a caveat: using a high-quality reconstruction algorithm will inevitably lead to higher distortion~\cite{blau2019rethinking}. Despite this, using PSC, the same compressed representation can be decoded using either a low-distortion or high perceptual quality approach with minimal additional computational cost. Specifically, we find that $\Pi$GDM~\cite{song2023pseudoinverse} produces images with highest photorealism, while DDRM~\cite{kawar2022denoising} leads to the lowest distortion. We present both restoration solutions as \textbf{PSC - Perception} and \textbf{PSC - Distortion} accordingly.

Figure~\ref{fig:imagenet-256-curves} presents the rate-distortion and rate-perception curves of PSC compared to several established methods: classic compression techniques like JPEG~\cite{wallace1991jpeg}, JPEG2000~\cite{skodras2001jpeg}, and BPG~\cite{bpg}. We also compare to neural compression methods, such as ELIC~\cite{he2022elic} and its diffusion-based derivative IPIC~\cite{xu2024idempotence}, as well as HiFiC~\cite{mentzer2020high}, a prominent GAN-based neural compression method. Distortion is measured by averaging the PSNR across different algorithms for a given compression rate. Image quality is quantified using FID~\cite{fid}, estimated on 50 random 128$\times$128 crops from each image (inspired by the evaluation in~\cite{mentzer2020high}), and compared to the same set of baselines. The graphs demonstrate that PSC achieves comparable performance, particularly at low BPP regimes, when considering both distortion and image quality. Figure~\ref{fig:imagenet-256} showcases qualitative image samples compressed using different algorithms at the same rate, further supporting our findings. Notably, PSC achieves exceptional image quality despite the fact that it does not require any compression-specific training.

\begin{figure*}[t]
\vskip 0.2in
  \begin{center}
  \centerline{\includegraphics[width=\textwidth]{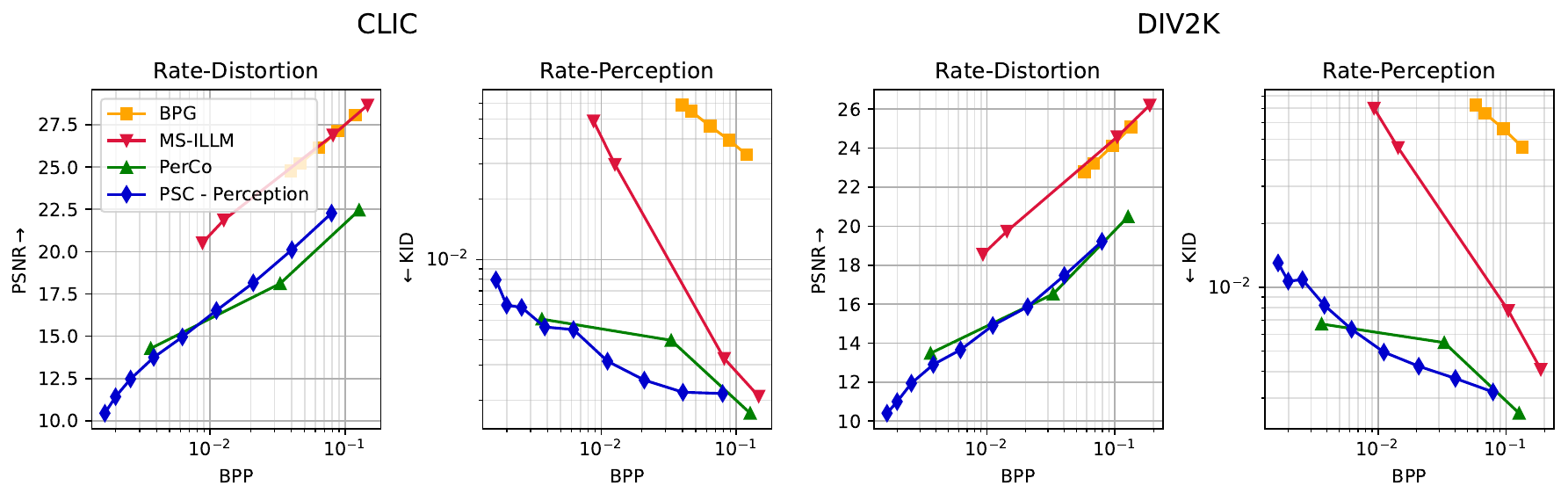}}
  \caption{\textbf{Rate-Distortion and Rate-Perception curves demonstrating PSC with latent-diffusion compared to similar methods.} Left: CLIC dataset, Right: DIV2K dataset. Distortion is measured as average PSNR, while Perception (photorealism) is measured by KID.}
    \label{fig:latent-curves}
  \end{center}
  \vskip -0.1in
\end{figure*}

\subsection{Latent PSC} 
Latent Text-to-Image diffusion models have gained popularity due to their ease-of-use and low computational requirements. These models employ a VAE~\cite{kingma2013auto} to conduct the diffusion process in a lower-dimensional latent space~\cite{vahdat2021score, latent_diffusion}. In this work we also explore the integration of PSC with Stable Diffusion~\cite{latent_diffusion}, a publicly available latent Text-to-Image diffusion model. 
This variant, named latent-PSC, operates in the latent space of the diffusion model. Both compression and decompression occur within this latent space, leveraging the model's VAE decoder to reconstruct the image from the decompressed latent representation. Additionally, we condition all posterior sampling steps on a textual description, which must be given along with the original image or inferred using an image captioning module~\cite{vinyals2016show, li2022blip, li2023blip}. The text prompt must be added to the compressed representation to avoid side-information. A detailed diagram of latent-PSC is presented in \Cref{app:implementation}.

Due to the use of the VAE decoder, we expect a significant drop in PSNR. Thus, we develop latent-PSC as an extension of \textbf{PSC - Perception}, maintaining high perceptual quality at low bit-rates. We find that the posterior sampler outlined in Nested Diffusion~\cite{Elata_2024_WACV} works best in this setting. We use images from the CLIC~\cite{CLIC2020} and DIV2K~\cite{div2k} to compare Latent-PSC to PerCo~\cite{careil2023towards}, a recent work which also utilizes latent diffusion for low-rate image compression. Using the same base diffusion model, image captioning model, and text compression as PerCo, we demonstrate in \Cref{fig:latent-examples} and \Cref{fig:latent-curves} that we are comparable in terms of both distortion (PSNR) and photorealism (KID~\cite{kid}) on all bit-rates despite not adding any training. We also compare to MS-ILLM~\cite{muckley2023illm}, another compression method that focuses on high perceptual quality, as well as BPG, as a classical baseline. While the results of MS-ILLM do not suffer from the VAE-induced drop in PSNR, they do not reach the image quality of PerCo or Latent-PSC, especially at very low rates.

\section{Related Work}

Diffusion models have enhanced classical compression algorithms by providing data-driven decompression for high-perceptual quality reconstruction~\cite{ghouse2023residual, saharia2022palette}. Several approaches implement zero-shot diffusion-based reconstruction~\cite{kawar2022jpeg, song2023pseudoinverse}, offering training-free decompression, but remain constrained to specific compression algorithms, which may be found lacking. Notably, recent work by~\citet{xu2024idempotence} attempts to utilize general diffusion-based posterior samplers ro increase the image quality using a leading neural compression method. While this work employs pre-trained diffusion-based posterior samplers similar to our method, they focus on traversing the RDP trade-off~\cite{blau2019rethinking} of existing neural compression schemes rather than developing a comprehensive compression solution.

Recent developments integrate neural compression with diffusion models for decompression. Some approaches employ separate~\cite{hoogeboom2023high} or joint~\cite{yang2024lossy} neural compression and diffusion training to create compact representations with conditional diffusion models for high-quality decompression. Specifically,~\citet{careil2023towards, relic2024lossy} utilize latent diffusion~\cite{latent_diffusion} and text-conditioned models for efficient training. Despite their promise, these methods require complex rate-specific training, limiting flexibility. While work by~\citet{gao2022flexible} addresses this through training-free post-hoc rate reconfiguration, they incur high computational costs and performance penalties.

The concept of leveraging pre-trained diffusion models for compression originated in the DDPM publication~\cite{ho2020denoising}, though it focused on theoretical compression limits rather than practical implementation. Similarly,~\citet{theis2022lossy} analyzes theoretical limits using reverse channel coding techniques~\cite{li2018strong}, but their implementation's high computational complexity and lack of public code prevents direct comparison with our approach.

\section{Limitation, Discussion and Conclusion}
While PSC introduces a novel approach to generative model-based compression, several limitations warrant discussion. The method's primary constraint is its substantial computational overhead due to iterative diffusion model sampling. This computational burden is intrinsically linked to the zero-shot posterior sampler's quality, though ongoing advances in diffusion models and posterior sampling techniques may significantly reduce this limitation.
The current implementation's simplified quantization strategy for measurements presents another constraint. Implementation of more sophisticated quantization methods could yield significant improvements in compression rates. Furthermore, PSC's restriction to linear measurements, imposed by both existing posterior sampler capabilities and the complexity of non-linear measurement optimization, indicates potential for enhancement through the exploration of non-linear measurements and corresponding inverse problem solvers.
Notwithstanding these limitations, PSC represents a significant advancement in zero-shot diffusion-based image compression through several distinct advantages. The method progressively acquires informative measurements to create compressed representations, with decompression faithfully reconstructing the original image by replicating the compression algorithm's steps. Its implementation simplicity, independence from training data, and cross-domain flexibility underscore its potential impact. Future developments promise to further advance this approach to both image compression and compression in general.

\clearpage\newpage

\section*{Acknowledgements}

This research was partially supported by the Council for Higher Education - Planning \& Budgeting Committee, Israel. This research was also partially supported by the Israel Science Foundation (grant no. 2318/22), by a gift from Elbit Systems, and by the Ollendorff Minerva Center, ECE faculty, Technion.
We would like to thank Hila Manor for her help in running evaluations, and Matan Kleiner and Sean Man for their helpful discussions and ideas.

\section*{Impact Statement}

This paper presents work whose goal is to advance the field of 
Machine Learning and Image Compression. There are many potential societal consequences of our work, none which we feel must be specifically highlighted here.

\bibliography{refs}
\bibliographystyle{icml2025}

%%%%%%%%%%%%%%%%%%%%%%%%%%%%%%%%%%%%%%%%%%%%%%%%%%%%%%%%%%%%%%%%%%%%%%%%%%%%%%%
%%%%%%%%%%%%%%%%%%%%%%%%%%%%%%%%%%%%%%%%%%%%%%%%%%%%%%%%%%%%%%%%%%%%%%%%%%%%%%%
% APPENDIX
%%%%%%%%%%%%%%%%%%%%%%%%%%%%%%%%%%%%%%%%%%%%%%%%%%%%%%%%%%%%%%%%%%%%%%%%%%%%%%%
%%%%%%%%%%%%%%%%%%%%%%%%%%%%%%%%%%%%%%%%%%%%%%%%%%%%%%%%%%%%%%%%%%%%%%%%%%%%%%%
\newpage
\appendix

\appendix
\onecolumn
\section{Implementation Details}
\label{app:implementation}
For the ImageNet experiment we used the unconditional diffusion model from~\citet{dhariwal2021diffusion} to apply PSC.  We used $25$ DDRM~\cite{kawar2022denoising} steps to generate $16$ samples, using $\eta = 1.0$ and $\eta_b = 0.0$. We added $12$ rows to $\mH$ in every iteration, and used matched the number of iterations to the desired rate. We restore the images using the same model with either $\Pi$GDM~\cite{song2023pseudoinverse} with 100 denoising steps and default hyperparameters for high perceptual quality restoration, or an average of 64 DDRM~\cite{kawar2022denoising} samples which where produced as detailed above for low-distortion restoration. We use the Range Encoder from constriction~\cite{bamler2022constriction} as an entropy encoder\footnote{\url{https://github.com/bamler-lab/constriction}} in all our experiments.

In the latent diffusion experiment we used float16 stable-diffusion-2-1-base\footnote{\url{https://huggingface.co/stabilityai/stable-diffusion-2-base}}~\cite{latent_diffusion} and $50$ DDRM steps, using the same hyperparameters. Due to the loss in PSNR due to using the VAE decoder, we focus Latent-PSC on high perceptual generation. We restore the images using the same model with the posterior sampler from Nested Diffusion~\cite{Elata_2024_WACV}, which applies 50 DDRM steps, each composed of 5 second order unconditioned diffusion sampling steps. We find this sampler to work better than $\Pi$GDM for this specific setup.

\begin{figure*}[t]
\vskip 0.2in
\begin{center}
  \centerline{\includegraphics[width=0.85\textwidth]{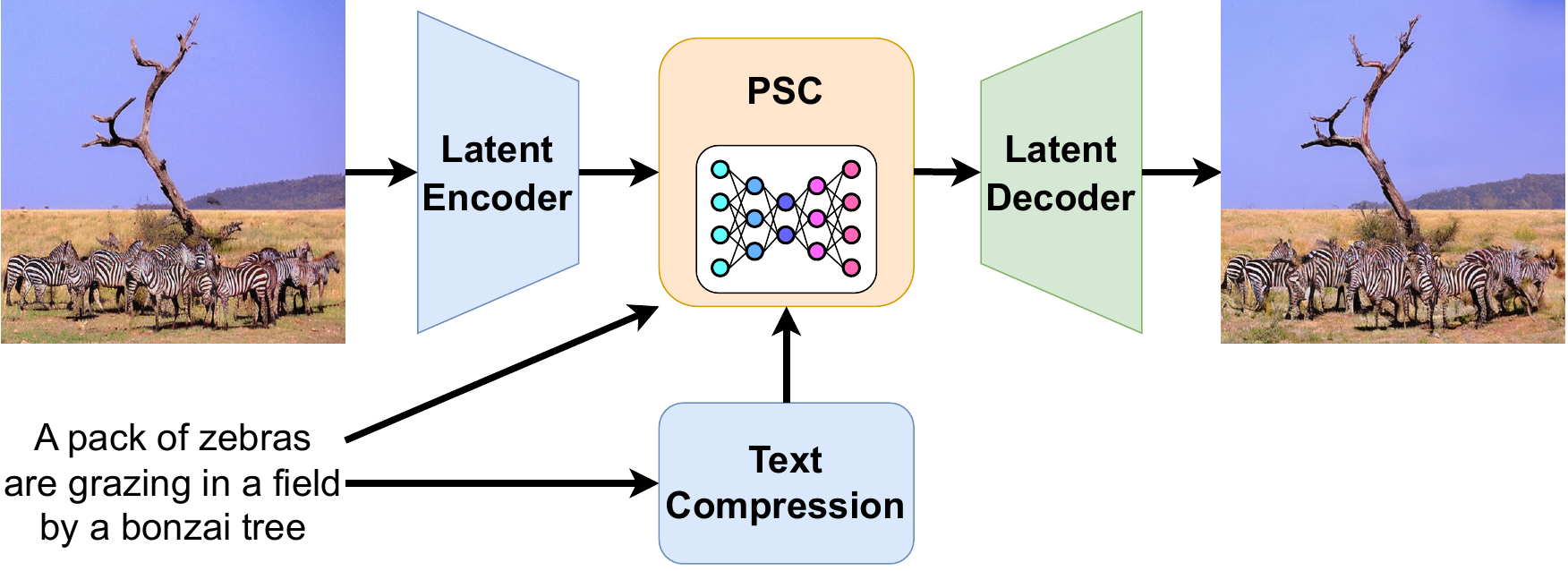}}  \caption{\textbf{Latent-PSC diagram:} Latent Text-to-Image diffusion models such as Stable Diffusion can be used for effective image compression with PSC. The latent representation is compressed using linear measurements. The textual prompt is used for conditioning the diffusion model in both the compression and decompression, and thus this text is also transmitted.}
    \label{fig:latent-diagram}
\end{center}
\vskip -0.2in
\end{figure*}

We used publicly available third party software for JPEG~\cite{wallace1991jpeg}, JPEG2000~\cite{skodras2001jpeg}, and BPG~\cite{bpg}. For HiFiC~\cite{mentzer2020high}, we trained our own model using the pytorch implementation publicly available on github\footnote{\url{https://github.com/Justin-Tan/high-fidelity-generative-compression}}. We trained the models using the default parameters for each rate, and pruned networks that failed to converge to the desired rate. To provide results for ELIC~\cite{he2022elic} and IPIC~\cite{xu2024idempotence} we used the official implementation for IPIC\footnote{\url{https://github.com/tongdaxu/Idempotence-and-Perceptual-Image-Compression}}. We used the Neural Compression\cite{muckley2021neuralcompression}\footnote{\url{https://github.com/facebookresearch/NeuralCompression/tree/main/projects/illm}} for MS-ILLM~\cite{muckley2023illm}
results, and the unofficial pytorch implementation\footnote{\url{https://github.com/Nikolai10/PerCo}} of PerCo~\cite{careil2023towards}. 

FID~\cite{fid} and KID~\cite{kid} is measured using Pytorch Fidelity~\footnote{\url{https://github.com/toshas/torch-fidelity}}. For FID on the $256\times256$ ImageNet images we used 50 random crops of size $128\times128$ inspired by~\citet{mentzer2020high}. For KID on experiments on the CLIC~\cite{CLIC2020} and DIV2K~\cite{div2k} validation datasets, we used $64\times 64$ patches from the $512\times512$ images, as a sufficiently large reference for high-quality images (The datasets are quite small).

The top image in \Cref{fig:latent-examples} is taken from the Kodak~\cite{kodak} Dataset.

\section{PSC Pseudo-Code}
\label{app:psuedo}
\begin{minipage}{\linewidth}
\begin{lstlisting}[language=Python]
from utilities import posterior_sampler, restoration_function, entropy_encode, entropy_decode

def SelectNewRows(H, y, r, shape, s=None):
    c, h, w = shape
    s = s if s is not None else (r * 4) // 3
    noise = torch.randn((s, c, h, w))
    samples = posterior_sampler(noise, H, y)
    samples = samples.reshape(s, -1)
    samples = samples - samples.mean(0, keepdim=True)
    new_rows = torch.linalg.svd(samples, full_matrices=False)[-1][:r]
    return new_rows

def PSC_compress(image, N, r)
    c, h, w = image.shape
    H = torch.zeros((0, c * h * w))         # Empty sensing matrix
    y = H @ image.reshape((-1, 1))          # Empty measurements
    compressed_representation = y.clone()
    
    for n in range(N):
        new_rows = SelectNewRows(H, y, r, (c, h, w)) 
        H = torch.cat([H, new_rows])
        y = torch.cat([y, new_rows @ image.reshape((-1, 1)])
        
        compressed_representation = y.to(torch.float8_e4m3fn) # Quantize
        y = compressed_representation.to(torch.float32)
        
    return entropy_encode(compressed_representation)

def PSC_decompress(compressed_representation, N, r)
    compressed_representation = entropy_decode(compressed_representation)
    c, h, w = image.shape
    H = torch.zeros((0, c * h * w))         # Empty sensing matrix
    y = H @ image.reshape((-1, 1))          # Empty measurements
    
    for n in range(N):
        new_rows = SelectNewRows(H, y, r, (c, h, w))
        H = torch.cat([H, new_rows])
        y = compressed_representation[:(n*r + r)].to(torch.float32)
        
    return restoration_function(H, y)

\end{lstlisting}
\end{minipage}

Our complete code will be published upon acceptance.

Latent-PSC follows the diagram in \Cref{fig:latent-diagram}, encoding the input image with the VAE encoder before compression and decoding the decompressed output with the VAE decoder. Also, as the diffusion model used is text-conditioned, the text prompt is given to the diffusion model inside \lstinline[language=Python]{posterior_sampler} and \lstinline[language=Python]{restoration_function}. The text is also begin compressed and appended to the compressed representation.

\section{Effect of measurement rank}
\label{app:rank}

\begin{figure}[t]
  \begin{center}
  \includegraphics[width=\textwidth]{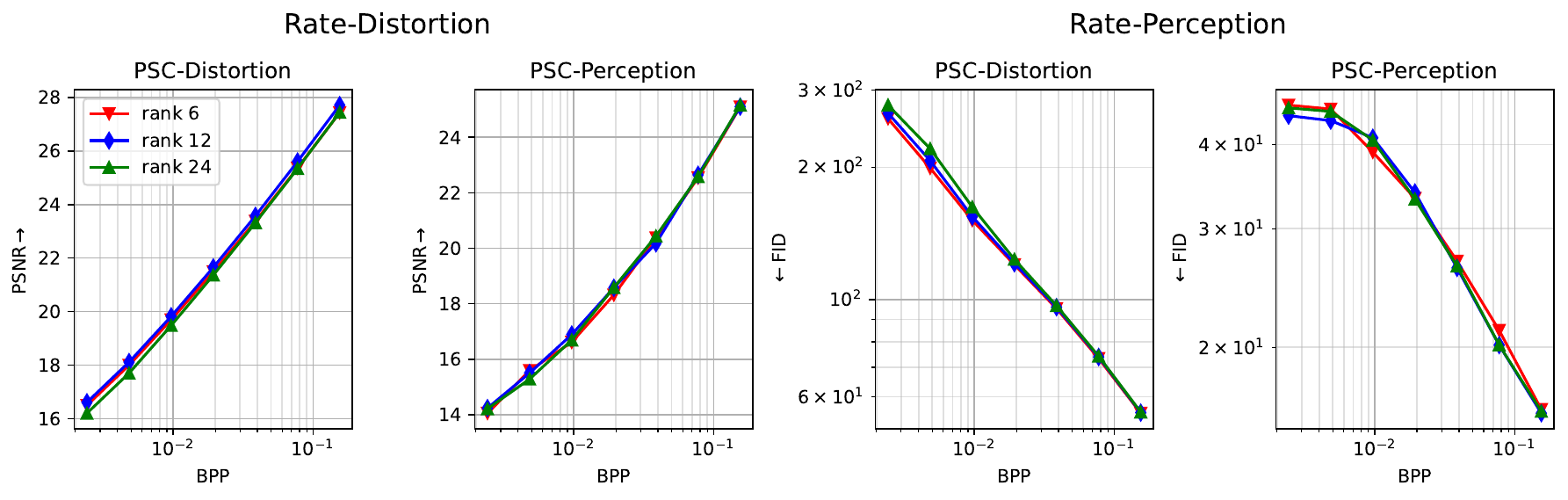}
  \caption{\textbf{Rate-Distortion (top) and Rate-Perception (bottom) curves for ImageNet256 compression, using PSC-Distortion (left) and PSC-Perception (right).} Distortion is measured as average PSNR of images for the same desired rate or specified compression quality, while Perception (image quality) is measured by FID.}
    \label{fig:imagenet-rank}
  \end{center}
\end{figure}

We repeat the ImageNet experiment with different values of the hyperparameter $r$, which determines how adaptive our algorithm would be. We modify the number of samples generated at each iteration $s$ accordingly to account for the rank required by the empirical covariance matrix. Based on the original implementation of AdaSense~\cite{elata2024adaptive}, we expect performance to improve the lower the value of $r$ is. In the results, demonstrated in \autoref{fig:imagenet-rank}, the variation of the rank seems to have only a marginal effect, even for low rates. We conclude that PSC is not sensitive to this parameter, and $r$ can be tuned according to the system's hardware (namely, maximum available batch size).

\section{Additional Latent-PSC ablations}

\begin{figure*}[t]
\vskip 0.2in
\begin{center}
  \centerline{\includegraphics[width=0.82\textwidth]{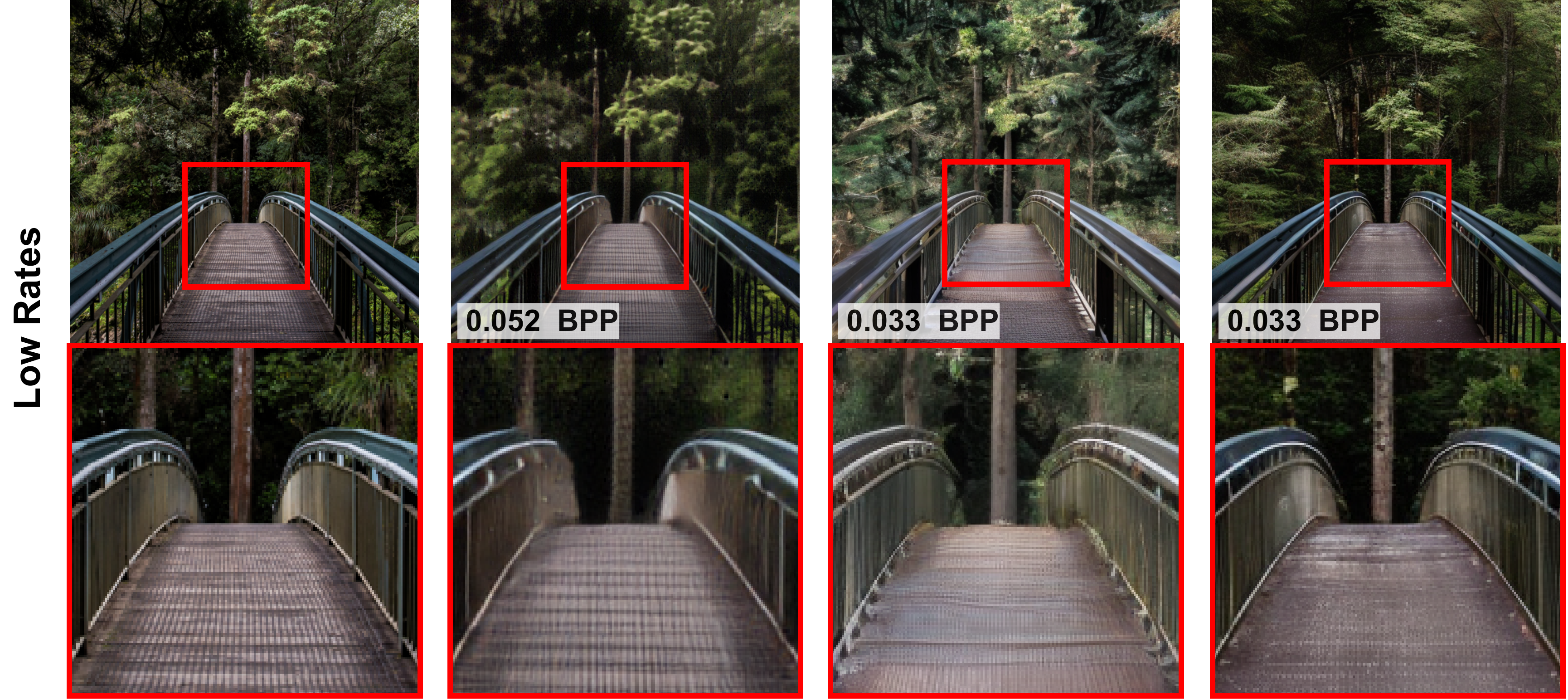}}
  \caption{\textbf{Additional comparison of Latent-PSC to leading methods.} Zoom-in view is shown below each image}
    \label{fig:additional-stable}
\end{center}
\vskip -0.2in
\end{figure*}

\begin{figure*}[t]
\vskip 0.2in
\begin{center}
  \centerline{\includegraphics[width=\textwidth]{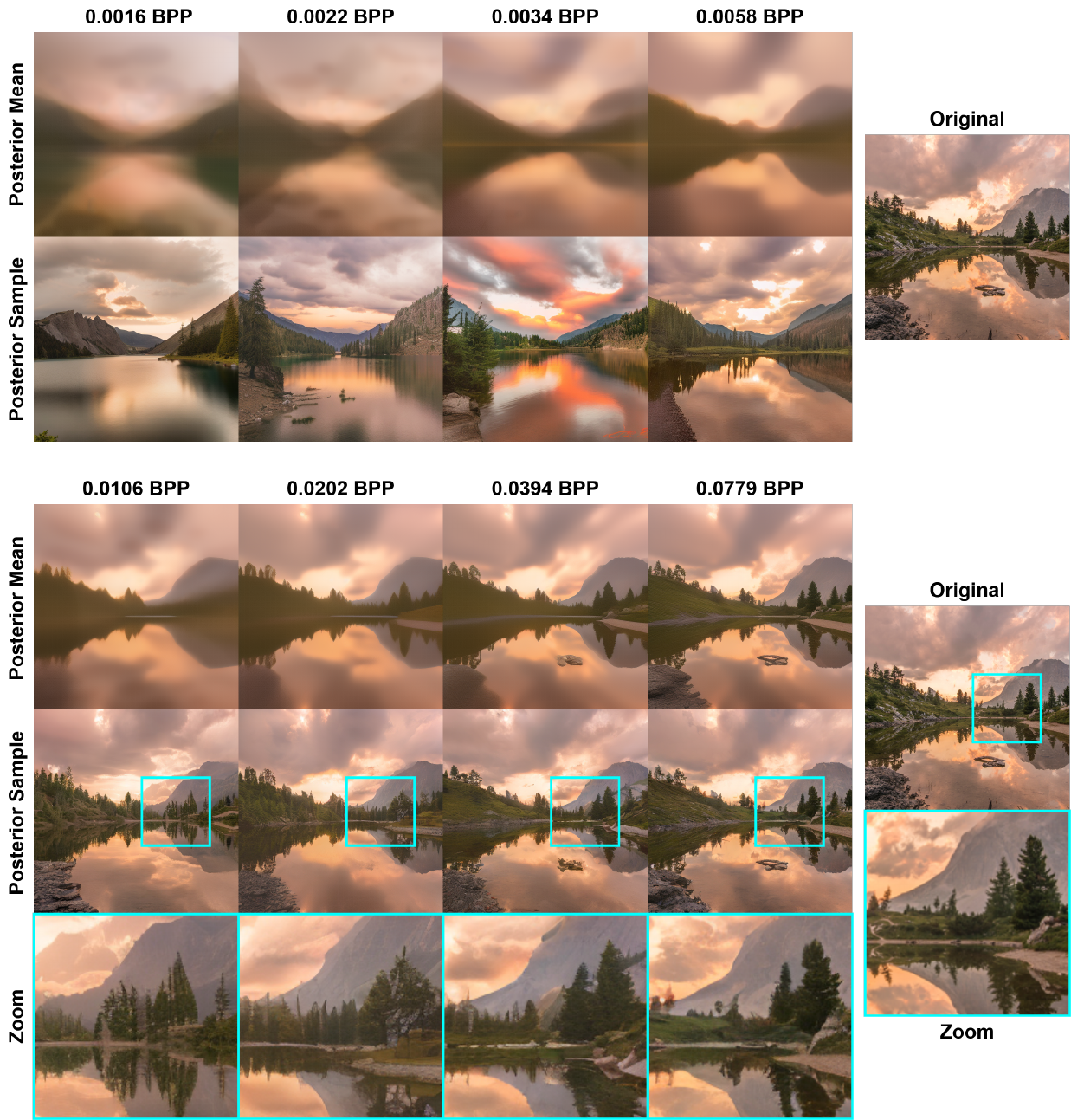}}
  \caption{\textbf{An example of the progression of PSC with growing bit-rates from left to right (both rows).} The posterior mean (top of each row) becomes sharper as more information about the original image is accumulated. The posterior samples also converge to the original while always being of high photorealism. A zoom-in view is shown below the bottom row of images to highlight fine differences}
    \label{fig:progressive}
\end{center}
\vskip -0.2in
\end{figure*}

\begin{figure*}[t!]
\vskip 0.2in
\begin{center}
  \centerline{\includegraphics[width=0.82\textwidth]{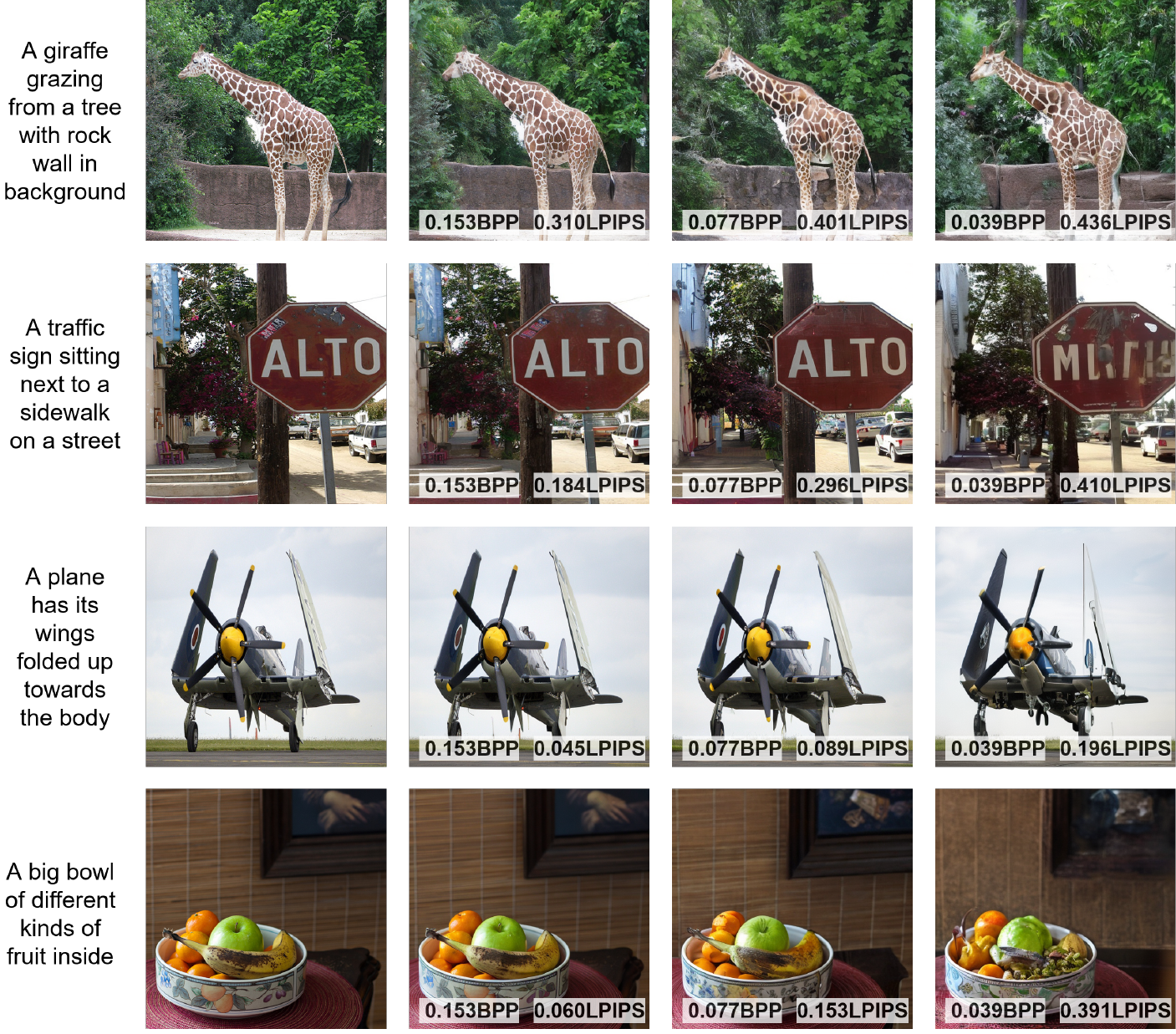}}
  \caption{\textbf{Qualitative examples of Latent-PSC with Stable Diffusion.} For each image and corresponding text, several results for different bit-rates are shown. BPP and LPIPS are reported.}
    \label{fig:stable}
\end{center}
\vskip -0.2in
\end{figure*}

Figure~\ref{fig:additional-stable} shows an additional comparison of Latent-PSC to PerCo~\cite{careil2023towards} and MS-ILLM~\cite{muckley2023illm}, similar to \Cref{fig:stable}. Figure~\ref{fig:progressive} demonstrates the progressive nature of PSC -- as more rows of $\mH$ are accumulated, the posterior distribution converges with the input. This comes at the cost of a larger compressed representation and higher bit-rates.

To quantify the effects of different rates and use of textual prompts, we evaluate Latent-PSC on $512\times512$ images from the MSCOCO~\cite{lin2014microsoft} dataset, which includes textual descriptions for each image.
We compress the textual description assuming 6 bits per character, with no entropy encoding. \autoref{fig:stable} shows decompressed samples using Latent-PSC with different rates, demonstrating good semantic similarity to the originals and high perceptual quality. 

\subsection{Effect of Caption on Latent-PSC}
\begin{figure*}[t!]
\vskip 0.2in
\begin{center}
  \centerline{\includegraphics[width=0.75\textwidth]{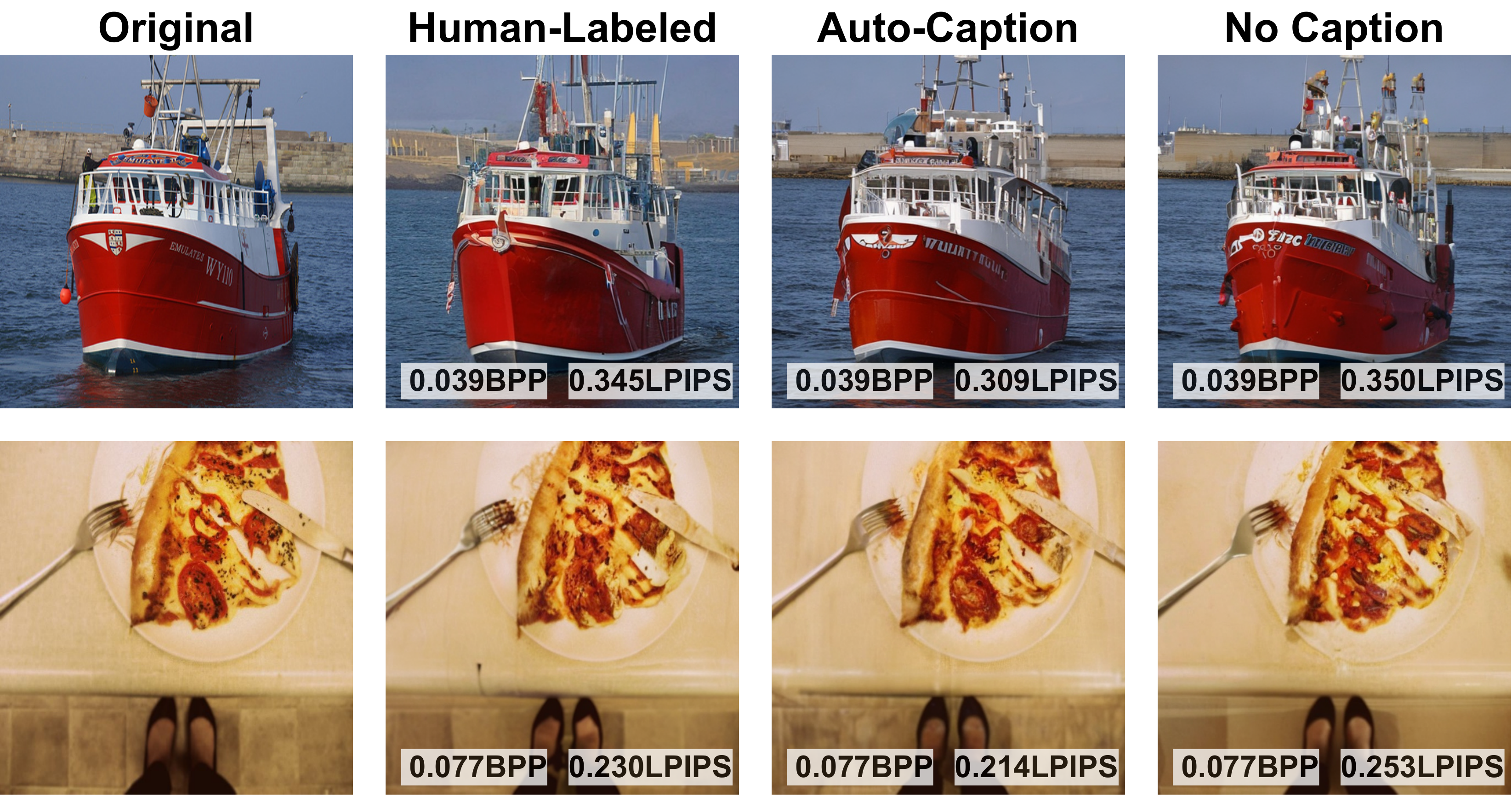}}
  \caption{\textbf{Qualitative examples of Latent-PSC with various prompt configurations.} For each image we compare compression results with human annotated textual description, auto-captioning using a model, and using no caption.}
    \label{fig:captioning}
\end{center}
\vskip -0.2in
\end{figure*}

\autoref{fig:captioning} illustrates the impact of using a captioning model to obtain the textual representation. In this experiment, the captions generated by BLIP~\cite{li2022blip} achieved comparable or superior results to human annotated description from the dataset. However, omitting the prompt causes some degradation of quality.

% \section{Image Specific Rate-Distortion}
% \label{app:image-specific}
% Below in \autoref{fig:image-specific-256-curves}, we present image-specific rate-distortion curves for the images displayed in \autoref{fig:imagenet-256}. These graphs provide additional evidence that the trends shown in \autoref{fig:imagenet-256-curves} is general to many images and not only to their avarage.

% \begin{figure}[h]
%   \begin{center}
%   \includegraphics[width=0.8 \textwidth]{figures/compress_imagenet_image_specific_256.pdf}
%   \caption{\textbf{Rate-Distortion curves for specific images from ImageNet256.} The images from \autoref{fig:imagenet-256} are used, numbered from top to bottom. Distortion is measured by PSNR of images.}
%     \label{fig:image-specific-256-curves}
%   \end{center}
% \end{figure}

\end{document}